# TOWARDS ANNOTATION OF TEXT WORLDS IN A LITERARY WORK

**Mikhalkova E. V.** (e.v.mikhalkova@utmn.ru),
**Protasov T.**, **Drozdova A.**, **Bashmakova A.**, **Gavin P.**

Tyumen State University, Tyumen, Russia

Literary texts are usually rich in meanings and their interpretation complicates corpus studies and automatic processing. There have been several attempts to create collections of literary texts with annotation of literary elements like the author's speech, characters, events, scenes etc. However, they resulted in small collections and standalone rules for annotation. The present article describes an experiment on lexical annotation of text worlds in a literary work and quantitative methods of their comparison. The experiment shows that for a well-agreed tag assignment annotation rules should be set much more strictly. However, if borders between text worlds and other elements are the result of a subjective interpretation, they should be modelled as fuzzy entities.

# К ВОПРОСУ О РАЗМЕТКЕ ТЕКСТОВЫХ МИРОВ В ЛИТЕРАТУРНОМ ПРОИЗВЕДЕНИИ

**Михалькова Е. В.** (e.v.mikhalkova@utmn.ru),
**Протасов Т.**, **Дроздова А.**, **Башмакова А.**, **Гэвин П.**

Тюменский государственный университет, Тюмень, Россия

Художественная литература обычно богата значениями и дает возможность для множества интерпретаций, которые усложняют корпусные исследования и автоматическую обработку. На данный момент попытки создать коллекции литературных текстов с размеченными литературными элементами, такими как речь автора, персонажи, события, сцены и т. д. не привели к созданию богатой коллекции или широкому распространению какого-либо набора правил для аннотации. В данной статье мы комбинируем литературный подход к интерпретации текста и теорию текстовых миров когнитивной лингвистики, чтобы изучить лексическую аннотацию текстовых миров в приключенческом романе («Вспомнить все», Ф. К. Дик). Эксперимент показывает, что объяснение литературных терминов и теории текстовых миров аннотаторам не является достаточным для хорошо согласованной разметки тегов, и некоторые правила должны быть установлены более строго. С другой стороны, если текстовые миры и их элементы должны отражать субъективное восприятие, тогда их нужно моделировать как нечеткие множества.

**Ключевые слова:** компьютерная лингвистика, когнитивная лингвистика, разметка текста, автоматическая обработка текста, теория текстовых миров, Филип Киндред Дик, Вспомнить все

## 1. Introduction

The Text World Theory (TWT), created in cognitive linguistics [5], [15], has recently gained interest of computer scientists as it provides a method for processing literary narrative or any other type of speech that needs to be interpreted from the point of view of its characters/speakers. For example, it helps to systematize stories of crime witnesses and compare details of crime scenes [8]. It can also help to compare characters of a novel, extract their speech from narration [11], etc. However, TWT still requires a better formalization which is especially difficult when it comes to a text world extraction.

In cognitive linguistics, a **text world** is a stretch of text perceived by a reader as a whole part of a narrative and characterized by a union of world-building elements, i.e. limited by a time period and space that holds all of its participants (or characters, in literary texts). The current research proposes several methods to compare text worlds across manually annotated texts. The paper is organized as follows. First, we describe an experiment on manual annotation of text worlds in Irina Utkina's Russian translation of the novel "We Can Remember It for You Wholesale" by Philip K. Dick [4][1]. Second, we suggest several quantitative methods to compare text worlds and their elements in six annotated texts. Finally, we conclude about possibilities of text world processing and outline perspectives for future work.

## 2. Related work

From the definition given above, the text world can be considered as the unity of characters, time and space. This unity is widely known in the literary theory as chronotope [1]. In TWT, whenever the characters change location or a significant amount of time passes, the world swithes to the next one. I.e. a text can be represented as a series of scenes that compose textual structure [12].

We know of the following software previously used to annotate text worlds and their elements:

- Worldbuilder [14].
- Concordance tools: CLIC [11] and WordSkew [2]. These are mainly used to study the author and characters' speech.

---

[1] Due to the copyright issues, contact the corresponding author to receive the collection of annotated texts.

- Visualisation tools: VUE [10].
- Tools for annotation of semantic roles: UCREL Semantic Annotation System, or USAS [6].

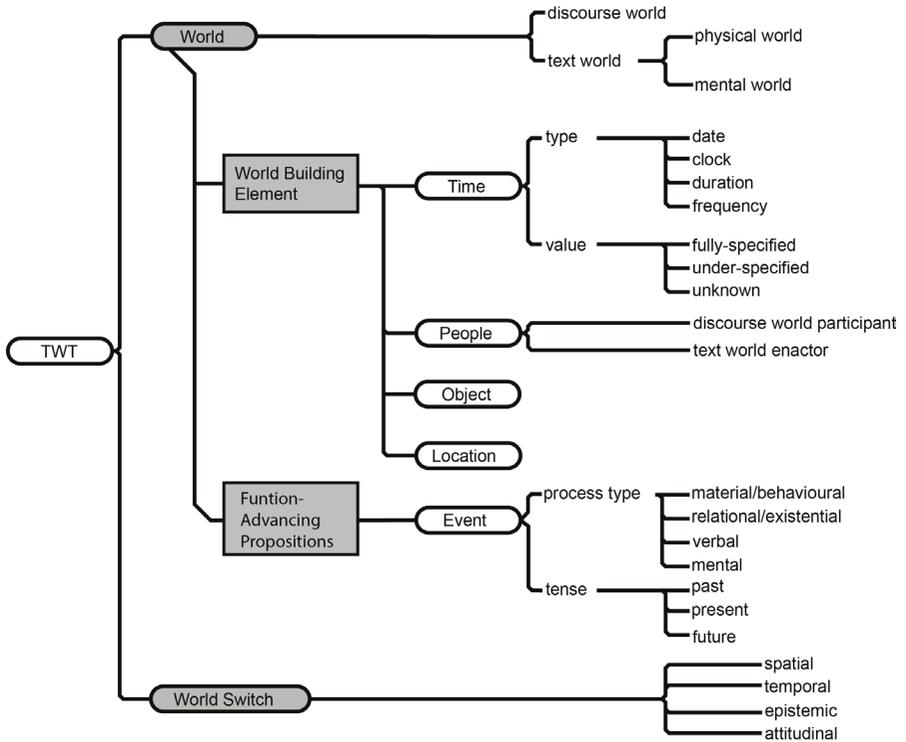

**Figure 1:** WorldBuilder annotation scheme

Of these systems, Worldbuilder has been designed particularly for annotation of text worlds. It is "a web front-end annotation tool for Text World analysis"[2] based on HTML5 and JavaScript. It processes texts with beforehand part-of-speech annotation, facilitates their manual annotation for TWT categories and shows descriptive statistics and visualisation of data. Such visualisation should help annotators to locate "worldbuilding elements" and text events. Unfortunately, WorldBuilder does not support large text files and has not been updated since around 2017. Later, creators of WorldBuilder used VUE instead [7], [8].

The original annotation scheme of WorldBuilder is given in **Fig. 1**.

---

2  http://www.viv-research.info/TWT/system/index.html

## 3. Experiment on manual annotation of text worlds

For our experiment on manual annotation of text worlds, we used XML format and the editor Sublime Text 3[3]. To process XML, we used the ElementTree XML API, version 1.0.1, for Python. Six experts with higher education in literature studies and linguistics annotated TWT elements in the novel "We Can Remember It for You Wholesale" by Philip K. Dick, an adventure narrative with distinguishable borders between text worlds. We have not proposed a set of strict rules for TWT-annotation, but conducted a seminar on discussion of its elements and how they are manifested in a literary text. For annotation, we used an abridged set of elements from Fig. 1:

- Place/location: <p[number]> Room D. </p[number]>.[4]
  ("Take <c1>me</c1> there, to <p6> Rekal, Incorporated </p6>").
- Time: <t>now</t>
- Character: <c[number]> Douglas Quail </c[number]>
- Switch: <s> descended </s>
- Text world: <T[number]> <c1> He </c1> awoke—and </T[number]>...

Table 1 contains data on the number of elements annotated in our collection. As can be seen from the Table, the number of annotated elements varies significantly. Moreover, it is hard to generalize about correlations within the Table, but for switches that signify a change of a text world. For example, Annotator 4 identified the maximum number of text worlds, switches and characters, and Annotator 6, who identified minimum of text worlds, tagged the maximum number of space and time indicators.

**Table 1:** Number of text elements annotated in the corpus

| Annotator | Text world | Switch | Characters | Space | Time |
|-----------|-----------|--------|-----------|-------|------|
| 1 | 20 | 17 | 28 | 21 | 61 |
| 2 | 12 | 9 | 45 | 34 | 41 |
| 3 | 14 | 18 | 37 | 41 | 56 |
| 4 | **56** | **72** | **48** | 43 | 41 |
| 5 | 22 | 22 | 43 | 37 | 77 |
| 6 | 8 | 28 | 23 | **44** | **92** |
| Median | 14, 20 | 18, 22 | 37, 43 | 37, 41 | 56, 61 |

## 4. Comparison of TW annotation

To compare text worlds across annotated texts, we discriminate outliers from Table 1 and take two texts where the number of text worlds is the closest to median: Text 1 with 20 text worlds and Text 3 with 14 text worlds. It is important that in our experiment the number of text worlds can be lower, than the number of extracts that relate to these worlds, because the characters can exit from a scene and return into it later in the narrative, e.g. when they leave a room and return soon afterwards. We will use the following metrics for the text world comparison.

---

[3] https://www.sublimetext.com/3

[4] The marker of "number" indicates a unique number in order of appearance of a TWT element.

## 4.1. Text worlds and switches

The metric we use to compare text worlds across annotated texts is Levenshtein distance from NLTK package for Python [3] which is "the number of characters that need to be substituted, inserted, or deleted, to transform" one string into the other. The transposition included in package parameters is forbidden, as we deal with two completely similar texts, but their extracts (substrings that we call text worlds) have different boundaries.

The metric is applied to two stretches of text from two parallel annotated texts marked as a text world $(T_i, T_j)$. When all possible pairs are measured, we find the minimum edit distance in the list of pairs $l_{min}$. The pair $(T_i, T_j)$ with $l_{min}$ is considered to be a match in annotation, i.e. the two extracts are considered to belong to one text world.

Table 2 illustrates Levenstein distance for the first 30 stretches of text.

**Table 2:** Levenshtein distance between text worlds in two annotated texts. 1/2 str. No.—number of a stretch of text marked as a text world, in the order of appearance, in the first/second annotated text

| 1 str. No. | $l_{min}$ | 2 str. No. | 1 str. No. | $l_{min}$ | 2 str. No. | 1 str. No. | $l_{min}$ | 2 str. No. |
|---|---|---|---|---|---|---|---|---|
| 1 | 0 | 1 | 11 | 5 | 8 | 21 | 29 | 18 |
| 2 | 0 | 2 | 12 | 218 | 11 | 22 | 5 | 20 |
| 3 | 6 | 3 | 13 | 0 | 37 | 23 | 21 | 19 |
| 4 | 0 | 4 | 14 | 27 | 12 | 24 | 0 | 20 |
| 5 | 7 | 5 | 15 | 4 | 4 | 25 | 6 | 21 |
| 6 | 327 | 32 | 16 | 105 | 14 | 26 | 0 | 22 |
| 7 | 6 | 4 | 17 | 4 | 4 | 27 | 22 | 23 |
| 8 | 172 | 9 | 18 | 329 | 16 | 28 | 4 | 37 |
| 9 | 437 | 41 | 19 | 5 | 45 | 29 | 215 | 25 |
| 10 | 0 | 7 | 20 | 158 | 8 | 30 | 264 | 27 |

It is of interest that in the beginning the two annotators defined text worlds almost unanimously. However, starting with stretch 6, differences built up. Fig. 2 demonstrates $l_{min}$ from Table 2, sorted in the rising order (blue circles), and, connected with them, the difference in stretch numbers given in absolute value (red circles). For example, stretch No. 4 in the first text is at distance 0 from stretch No. 4 in the second text. Hence, their difference in stretch numbers is also 0. However, stretch 5 is at distance 327 from the corresponding stretch, and the ordinal number of the latter is 32, which means that their difference is $32 - 5 = 27$.

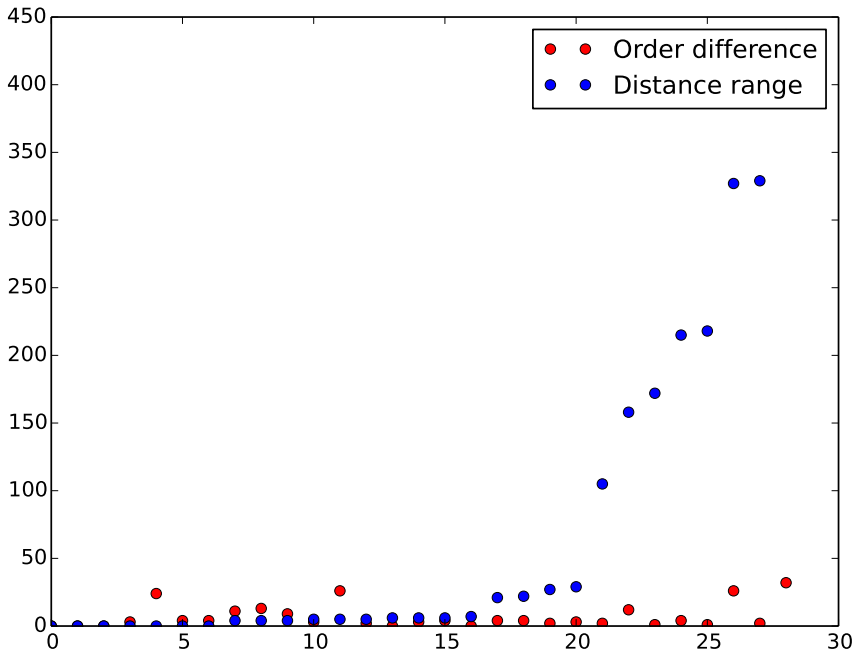

**Figure 2:** Differences in $l_{min}$

Mismatches between two worlds can be of two types. The first one is when a stretch annotated in the first text as a separate TW highly resembles some stretch from the second TW, but they are very far from each other according to their position in the text. Second, when the two stretches are quite near in the text, but they do not match each other in characters. For example, when one annotator split some part of a text into more TWs than another annotator.

As for switches between TWs, they reflect the annotators' disagreement as follows. There were only two places in the text where the number of annotators that agreed about a change of TW was five. One switch was "* * *" and the other—the verb "Приехав". Also, four annotators agreed on marking verbs "выглянул" and "захотел" as switches; three—on verbs "вернулся", "приехали", "Выйдя". Most switches are annotated by just one or two annotators and consist of one word. When we POS-tagged them with PyMystem3[5], it occurred that they are, or contain, mainly verbs (134). Also, among them there are 41 noun, 33 adjectives, 30 punctuation symbols like "—", 16 other types. In four cases, the annotators missed a switch between TWs.

## 4.2. Text world elements

**Table 3** illustrates the choice of tags for eight text world elements that were among the first to appear in annotation. The main character of the novel, Douglas

---

Quail, and the most mentioned place, Mars, that were both introduced in the very beginning, got similar tags. However, with every new line differences built up. For example, half of the annotators decided that valleys on Mars are a different location than the planet as a whole. Consequently, the annotators' agreement calculated by homogeneity measures like Fleiss' kappa would be very low. For now, let us assume that assigning a tag is a probabilistic judgement, i.e. every text world component has a certain probability among readers to be judged as a character, location, etc. For example, according to our experiment, the main character is defined as such by 100% of annotators, and probably it will be so if we ask more experts. However, parts of locations (like valleys on Mars, or a bedroom and kitchen in an apartment) have an approx. 50% chance to be considered a separate location.

**Table 3:** Comparison of tags assigned to similar elements

| Text elements | 1 | 2 | 3 | 4 | 5 | 6 |
|---|---|---|---|---|---|---|
| D. Quail | c1 | c1 | c1 | c1 | c1 | c1 |
| Mars | p1 | p1 | p1 | p1 | p1 | p1 |
| Government agents | c2 | c2 | c2 | c2 | c2 | — |
| high officials | c3 | c3 | c3 | c3 | c3 | — |
| a clerk | — | — | — | c4 | — | — |
| Kirsten | c4 | c4 | c4 | c5 | c4 | c2 |
| Valleys on Mars | — | p2 | — | p2 | — | p2 |
| Quail's home | p2 | p5 | p2 | — | — | — |

In the current research, we will view elements of a text world as multitudes of lexical forms that were tagged with a similar tag and compare them across annotated texts with Jaccard similarity coefficient $J(A, B)$ [9]:

$$J(A, B) = \frac{|A \cap B|}{|A| + |B| - |A \cap B|}.$$

We apply $J$ pairwise to a list of all possible pairs of annotated texts $(t_i, t_j)$ and calculate $\bar{J}$ to see, which of the tagged elements are most similar:

$$\bar{J} = \frac{\sum J(t_i, t_j)}{N},$$

where $N$ is the number of possible pairs.

We expect that elements with low similarity require a more elaborated rule for annotation. Within the limits of the current article, we will consider only characters and places. Table 4 demonstrates elements with the highest similarity coefficient.

The reader might notice, especially if they are familiar with the novel, that elements in Table 4 can be divided into two types: those that appeared in short scenes and did not get an extended description (e.g. parts of an interior) and those that show up many times (e.g. the main characters). Between these two extremes, there is a number of elements that confused annotators and got $\bar{J} < 0.7$. Among these are groups that are mentioned hypothetically, without much reference to their activity and, hence, influence on events: *women, children, people, somebody, armed men, workers, clerks, everybody*. Some groups that are in close proximity to the main characters are defined more unanimously, e.g. passers-by ($\bar{J} = 0.62$) in the street where D. Quail is walking or Government agents whom he can meet at Interplan ($\bar{J} = 0.6$). Another case is large locations and their parts. Larger places like Mars, an appartment (con-apt), a city (New-York, Chicago) can be split into smaller ones (Mars' moons $\bar{J} = 0.45$, craters $\bar{J} = 0.03$, volcanos). This is especially confusing when the action does not take place in any of these sub-locations. Also, in case of organizations and moving mechanisms, like Rekal Incorporated ($\bar{J} = 0.35$), cars, spaceships etc. that can be considered a place or character, annotators did not agree on the *type* of TW element. In a strict set of rules for annotation, these cases, probably, require a separate guideline.

**Table 4:** Similarity of different TW elements across pairs of annotated texts: 1—Luna, 2—Another star system, 3—Douglas Quail, 4—Robot-driver, 5—Political organization, 6—Kirsten, 7—McClane, 8—Aliens, 9—Shirley, 10—High officials, 11—Elderly Interplan psychiatrist, 12—Policeman, 13—Mars, 14—Chamber behind office, 15—Park, 16—UNO General Secretary

| Elem. | (1,2) | (1,3) | (1,4) | (1,5) | (2,3) | (2,4) | (2,5) | (3,4) | (3,5) | (4,5) | $\bar{J}$ |
|---|---|---|---|---|---|---|---|---|---|---|---|
| 1 | 1.0 | 1.0 | 1.0 | 1.0 | 1.0 | 1.0 | 1.0 | 1.0 | 1.0 | 1.0 | 1.0 |
| 2 | 0.89 | 0.89 | 0.89 | 1.0 | 1.0 | 1.0 | 0.89 | 1.0 | 0.89 | 0.89 | 0.93 |
| 3 | 0.93 | 0.93 | 0.91 | 0.92 | 0.93 | 0.89 | 0.9 | 0.9 | 0.92 | 0.89 | 0.91 |
| 4 | 0.75 | 0.75 | 0.75 | 0.75 | 1.0 | 1.0 | 1.0 | 1.0 | 1.0 | 1.0 | 0.9 |
| 5 | 1.0 | 0.75 | 1.0 | 1.0 | 0.75 | 1.0 | 1.0 | 0.75 | 0.75 | 1.0 | 0.9 |
| 6 | 0.94 | 0.96 | 0.86 | 0.9 | 0.98 | 0.81 | 0.88 | 0.83 | 0.87 | 0.8 | 0.88 |
| 7 | 0.86 | 0.91 | 0.83 | 0.89 | 0.86 | 0.85 | 0.86 | 0.81 | 0.86 | 0.84 | 0.86 |
| 8 | 0.88 | 0.89 | 0.8 | 1.0 | 0.78 | 0.83 | 0.88 | 0.77 | 0.89 | 0.8 | 0.85 |
| 9 | 0.88 | 0.85 | 0.79 | 0.89 | 0.97 | 0.9 | 0.77 | 0.87 | 0.79 | 0.74 | 0.84 |
| 10 | 1.0 | 0.5 | 1.0 | 1.0 | 0.5 | 1.0 | 1.0 | 0.5 | 0.5 | 1.0 | 0.8 |
| 11 | 0.9 | 0.7 | 0.9 | 0.64 | 0.78 | 1.0 | 0.7 | 0.78 | 0.88 | 0.7 | 0.8 |
| 12 | 0.61 | 0.95 | 0.91 | 0.95 | 0.6 | 0.61 | 0.58 | 0.87 | 0.91 | 0.87 | 0.79 |
| 13 | 0.74 | 0.81 | 0.73 | 0.77 | 0.74 | 0.79 | 0.91 | 0.66 | 0.71 | 0.83 | 0.77 |
| 14 | 1.0 | 0.33 | 1.0 | 1.0 | 0.33 | 1.0 | 1.0 | 0.33 | 0.33 | 1.0 | 0.73 |
| 15 | 1.0 | 0.33 | 1.0 | 1.0 | 0.33 | 1.0 | 1.0 | 0.33 | 0.33 | 1.0 | 0.73 |
| 16 | 1.0 | 0.56 | 0.67 | 1.0 | 0.56 | 0.67 | 1.0 | 0.38 | 0.56 | 0.67 | 0.7 |

## 5. Conclusion

Concerning the studied text worlds and their elements, their processing seems to be a combination of existing NLP-tasks. In our experiment, switches, signifying borders between text worlds, tend to be verbs (*morphological parsing*) of activity. Usually, they are syntactically bound (*sytntactical parsing*) to the doer of the action who is one of the main characters of the novel and is often mentioned within a text world. As for TW-elements, characters and locations are objects of NER-parsing and anaphoric resolution. Time passage can be detected semantically or in grammatical forms. However, before this is possible, we need to create an extended annotated corpus of literary texts.

So far we have described how we applied literary theory and TWT in annotation of a literary text. Our experiment shows that theoretical understanding of literary terms like *chronotope* and *character* and terminology of the TWT is not enough for a well-agreed annotation of world-building elements. We have discussed some of the problematic issues that require strict rules. However, more can be inferred from our collection of annotated texts. Creating such a list and an annotated corpus of literary texts is our next step. However, in conclusion, we would also like to outline broader perspectives for future work.

First of all, defining such elements as "the main character", "supporting character" etc. can become algorithmic based on the calculation of mentions that these elements have and in how many text worlds they appear. These calculations can serve as verification of expert analysis. Second, features extracted from TW-annotation can be used for quantitative analysis of genres, styles, etc. Also, these features can be the basis for comparison of the original text and its translation, fanfiction, sequel, etc. Third, as concerns translation, tags signifying borders between text worlds might improve performance of machine and computer-assisted translation systems (at present, segmentation in computer-assisted translation is based on punctuation that signals the end of a sentence [13]). Also, dialogue systems might acquire ability to understand a situation, context of conversation, and focus on its participants. However, the last two tasks require that TW-annotation becomes automatic.

## Acknowledgements

The reported study was funded by RFBR according to the research project No. 18-312-00127.

We would also like to thank Dmitry Bashmakov for his help with the annotation of the text.